\documentclass{article}
\usepackage{array}
\usepackage{booktabs}
\usepackage{float}
\usepackage{microtype}
\usepackage{graphicx}
\usepackage{subcaption}
\usepackage{booktabs} 
\usepackage{multirow} 
\usepackage{hyperref}

\usepackage{wrapfig}  
\usepackage{overpic}
\usepackage[numbers]{natbib}

\usepackage[preprint]{neurips_2026}

\usepackage[utf8]{inputenc} 
\usepackage[T1]{fontenc}    
\usepackage{hyperref}       
\usepackage{url}            
\usepackage{booktabs}       
\usepackage{amsfonts}       
\usepackage{nicefrac}       
\usepackage{microtype}      
\usepackage{xcolor}         
\usepackage{algorithm}
\usepackage{algorithmic}

\usepackage{amsmath}
\usepackage{amssymb}
\usepackage{mathtools}
\usepackage{amsthm}
\usepackage{graphicx}
\usepackage{multirow}
\usepackage[capitalize,noabbrev]{cleveref}
\usepackage{fontawesome5}
\usepackage[most]{tcolorbox}
\theoremstyle{plain}

\theoremstyle{definition}

\theoremstyle{remark}

\usepackage[textsize=tiny]{todonotes}
\title{HyMem: Hybrid Memory Architecture with Dynamic Retrieval Scheduling}

\author{%
  Xiaochen Zhao$^{1}$ \quad
  Kaikai Wang$^{2}$ \quad
  Xiaowen Zhang$^{2}$ \quad
  Chen Yao$^{2,*}$ \quad
  Aili Wang$^{1,*}$ \\  
  $^1$ ZJU-UIUC Institute, Zhejiang University \\
  $^2$ Ant Group \\
  \texttt{chen.yaoc@antgroup.com, ailiwang@intl.zju.edu.cn}
}

\begin{document}

\maketitle
{\renewcommand\thefootnote{*}\footnotetext{Corresponding authors}}

\begin{abstract}
Large language model (LLM) agents demonstrate strong performance in short-text contexts but often underperform in extended dialogues due to inefficient memory management. Existing approaches face a fundamental trade-off between efficiency and effectiveness: memory compression risks losing critical details required for complex reasoning, while retaining raw text introduces unnecessary computational overhead for simple queries. The crux lies in the limitations of monolithic memory representations and static retrieval mechanisms, which fail to emulate the flexible and proactive memory scheduling capabilities observed in humans, thus struggling to adapt to diverse problem scenarios. Inspired by the principle of cognitive economy, we propose HyMem, a hybrid memory architecture that enables dynamic on-demand scheduling through multi-granular memory representations. HyMem adopts a dual-granular storage scheme paired with a dynamic two-tier retrieval system: a lightweight module constructs summary-level context for efficient response generation, while an LLM-based deep module is selectively activated only for complex queries, augmented by a reflection mechanism for iterative reasoning refinement. Experiments show that HyMem achieves strong performance on both the LOCOMO and LongMemEval benchmarks, outperforming full-context while reducing computational cost by 92.6\%, establishing a state-of-the-art balance between efficiency and performance in long-term memory management. Our code is available at \href{https://github.com/xiaochenzhao-svg/HyMem}{https://github.com/xiaochenzhao-svg/HyMem}.
\end{abstract}

\section{Introduction}


Large language models (LLMs) inherently operate in a stateless manner, which limits their reasoning capabilities in long-context conversation scenarios \cite{chen2024large,li2024personal,tseng2024two}. Recent research \cite{chhikara2025mem0,liu2024agentlite,packer2023memgpt,roucher2025smolagents,zhong2024memorybank} has primarily focused on designing memory systems to enable efficient storage and retrieval of information for LLM-based agents. These approaches typically treat memory as external knowledge stored in vector databases \cite{edge2024local,lewis2020retrieval,shi2024commands}. To mimic organizational principles of human memory, works such as \cite{chhikara2025mem0,li2025cam,xu2025mem,edge2024local,han2024retrieval} extract entities and construct relational structures among memory objects. Concurrently, several studies \cite{wang2025mirix,wang2025mem,hu2025memory} have proposed multi-category memory segmentation and batch retrieval schemes, thereby providing fundamental support for personalized user memory needs.
However, existing methods face two core challenges. First, a single granularity of memory storage struggles to achieve an optimal balance between efficiency and effectiveness. To reduce computational overhead in long dialogues, studies like \cite{fang2025lightmem, chhikara2025mem0, packer2023memgpt, pan2024llmlingua} attempt to compress or distill raw text into lightweight memory libraries. Yet, such compression often leads to loss of critical details, resulting in irreversible "forgetting" phenomena when handling complex or detail-dependent queries \cite{yang2026beyond, tan2025prospect}. Conversely, storing raw text minimizes information loss but introduces unnecessary computational burden for simple queries. Second, static memory retrieval strategies fail to effectively emulate the dynamic, hierarchical recall patterns exhibited by humans in complex tasks. These limitations collectively constrain the flexibility and reliability of language agents in long-dialogue applications.

In contrast, the human memory system demonstrates notable cognitive economy: for routine queries, the brain prioritizes low-cost summary retrieval to quickly grasp event outlines; deeper logical reasoning and causal analysis are activated only when tasks require detailed support. This hierarchical, adaptive memory scheduling mechanism efficiently integrates memories at different levels, offering valuable inspiration for constructing LLM memory systems that reconcile efficiency with expressive power.

To address these challenges, we propose HyMem, a hybrid memory architecture integrated with a dynamic retrieval mechanism, designed to flexibly adapt to diverse QA scenarios. At the storage level, the architecture partitions raw dialogues into event units and constructs a dual-granularity storage structure comprising summary-level (Level-1 memory) and raw-text-level (Level-2 memory). During inference, HyMem adopts a dynamic retrieval strategy guided by cognitive economy: it first employs lightweight matching to locate Level-1 memory for rapid context construction; then, based on query complexity, it decides whether to activate Level-2 memory for finer-grained information. Within the deep retrieval module, the system performs coarse-grained recall to narrow candidate scope, followed by LLM-powered event-query relevance analysis for precise context localization. Additionally, a reflection module assesses answer completeness and triggers multi-round query refinement when necessary, enhancing the robustness and thoroughness of reasoning.
Our main contributions are:
\begin{itemize}
    \item We propose HyMem, a hybrid memory architecture grounded in the principle of cognitive economy. Through dual-granularity memory storage and dynamic on-demand scheduling, it achieves an effective balance between memory management efficiency and reasoning performance for large-scale language agents in long dialogues.
    \item We introduce an LLM-guided self-retrieval mechanism that dynamically activates deep memory access, combined with an LLM self-reflection process, enabling precise information localization and efficient, reliable iterative reasoning in complex query scenarios.
    \item We conduct extensive experiments on the LoCoMo and LongMemEval benchmarks. Results demonstrate that our approach yields significant advantages in answer flexibility and reliability across diverse scenarios.
\end{itemize}
\begin{figure*}[t] 
  \centering
  \includegraphics[width=\textwidth]{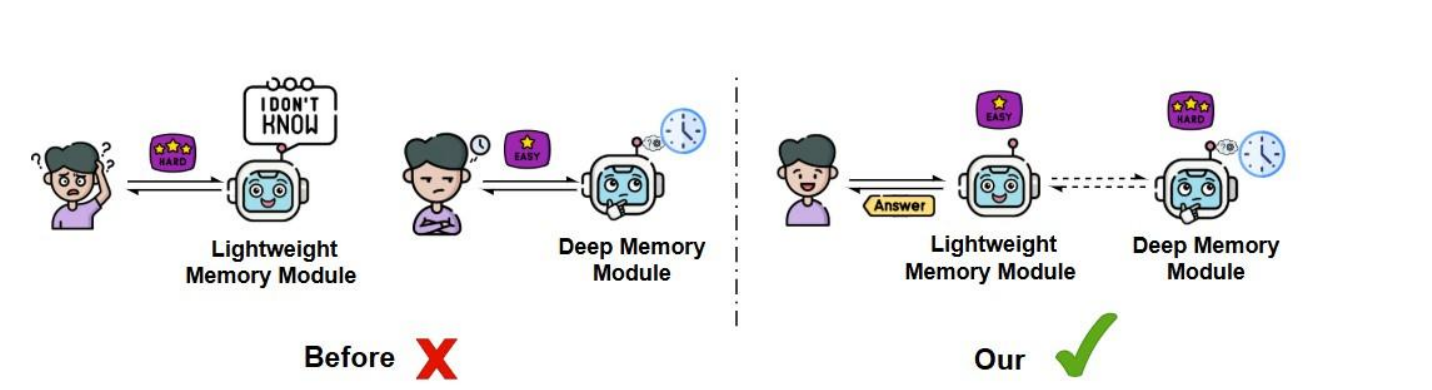}
  \caption{Conventional lightweight methods struggle with complex tasks, while sophisticated approaches incur high overhead for simple queries. In contrast, our HyMem dynamically allocates memory resources based on task demands, achieving dual optimization of performance and efficiency.}
  \label{fig:figure1}
\end{figure*}

\section{Related Work}
\subsection{Memory for LLM Agents}
To alleviate the inherent limitations of large language models (LLMs) in multi-turn long-context dialogues, a considerable body of recent research has focused on enhancing agents’ memory and adaptive evolution during interactions. Techniques such as \cite{munkhdalai2024leave,bulatov2022recurrent,su2024roformer,chen2023extending,an2024training,liu2023scaling,xiong2024effective,wei2025mlp,wutokmem,zhang2025memgen} inject additional signals during training or modify model architectures, aiming to improve retention and reasoning over distant information. However, these approaches generally require retraining the base model, incurring substantial computational and engineering expenses, and are incompatible with powerful closed-source models like GPT-4.

In contrast, retrieval-augmented generation (RAG) methods based on external knowledge bases have become a mainstream, efficient, and flexible solution \cite{hu2025memory,huang2026rethinking,jin2025flashrag,wu2025composerag,guo2024lightrag,lee2024planrag,asai2024self}. For example, \cite{packer2023memgpt} adopts a hierarchical memory structure and context paging, leveraging a controller to manage prioritized reading and writing to external vector databases, thereby achieving persistent long-term memory and dynamic injection. \cite{li2025memos} transforms the interaction process into an agent-oriented "memory operating system," which automatically extracts and stores user-relevant information. Approaches such as \cite{li2025cam,xu2025mem,jimenez2024hipporag,edge2024local,xia2025experience,anokhin2024arigraph,wu2025sgmem,rasmussen2501zep} draw inspiration from knowledge graphs to organize memories by extracting entities and constructing relationships among memory objects. Meanwhile, \cite{xu2025mem} proposes a self-updating paradigm, encompassing memory writing, retrieval, updating, and forgetting. Methods in \cite{wang2025mirix,wang2025mem,hu2025memory} introduce multiple structured memory modules and autonomous routing, enabling more fine-grained and personalized long-term memory.

Nevertheless, during the inference phase, prevailing approaches still predominantly rely on fixed memory retrieval workflows—such as those dependent on static vector representations and similarity search—for context construction. These methods exhibit significant limitations when confronted with compositional complex reasoning tasks, presenting a critical bottleneck that restricts agent interaction capabilities in long-context scenarios. To address this issue, our hybrid memory framework, HyMem, adaptively responds to query difficulty, empowering the LLM to proactively retrieve information necessary for complex questions rather than relying solely on vector embeddings. Furthermore, the framework incorporates a reflection mechanism to mitigate hallucination issues arising from unimodal retrieval in challenging scenarios.

\begin{wrapfigure}{r}{0.45\textwidth}
  \centering
  \vspace{-15pt}
\includegraphics[width=0.45\textwidth]{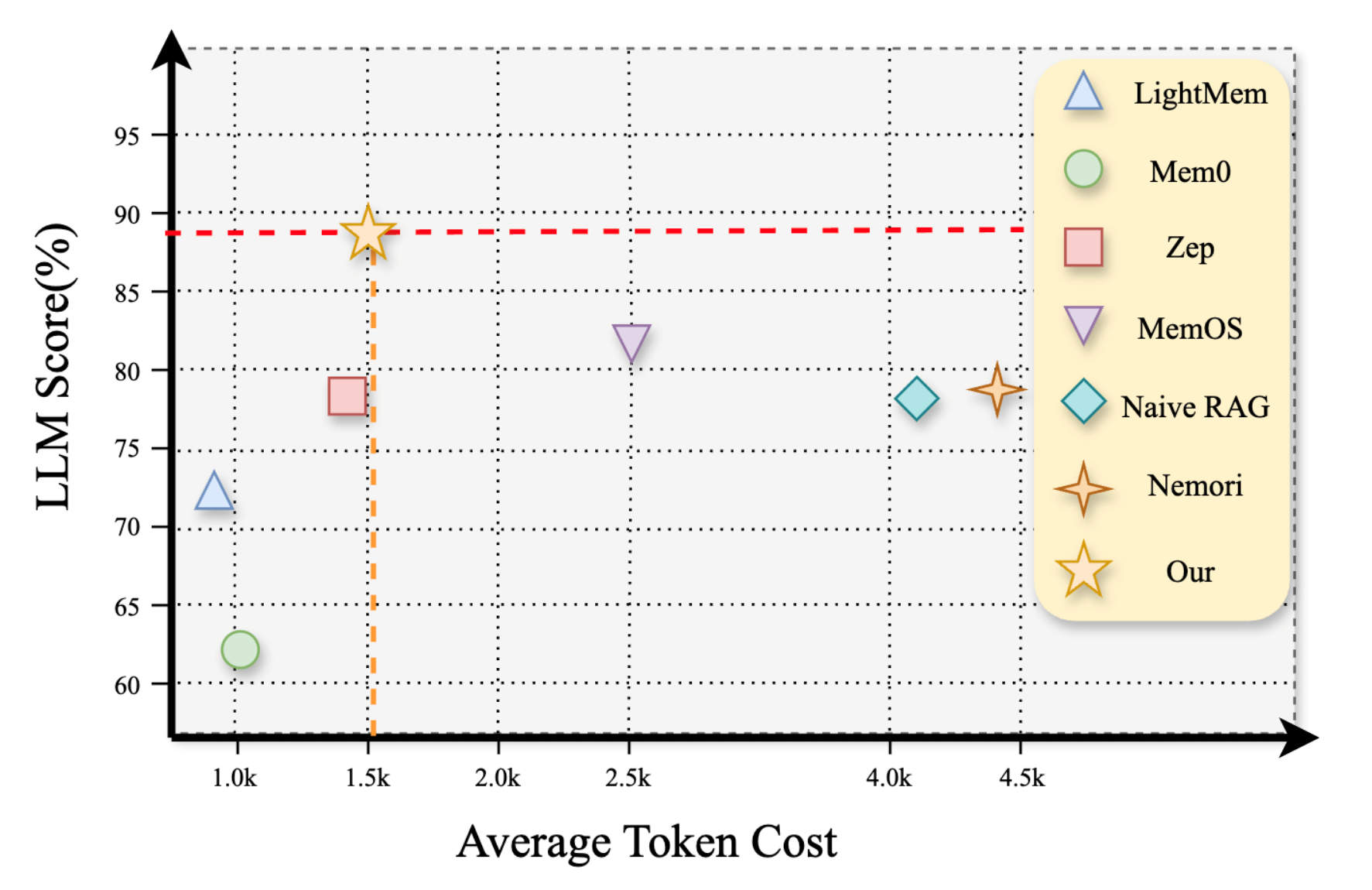}
  \caption{Our approach demonstrates superior efficiency, achieving the best balance between performance and computational cost on the LOCOMO benchmark.}
  \label{fig:figure4}
  \vspace{-15pt}
\end{wrapfigure}

\subsection{Context Management.}
Another important challenge is the rapid increase in token consumption as dialogue context expands. \cite{liu2026simplemem,fang2025lightmem} introduces token filtering and memory pre-compression to reduce retrieval and generation costs while preserving key information. \cite{xu2025mem} further compresses long conversations into discrete "atomic notes," effectively reducing token redundancy. \cite{chhikara2025mem0} extracts factual knowledge from dialogues and integrates it into user profiles, rather than storing entire conversation logs. \cite{wang2024memoryllm} compresses historical information into latent states to form a memory pool. However, such compression inevitably leads to permanent information loss. To address this, \cite{yan2025general,tan2025prospect} utilize only raw text to construct context, aiming to avoid information loss induced by memory compression. Recent studies such as \cite{yu2025memagent,zhou2025mem1,ye2025agentfold} have effectively mitigated the “forgetting” problem in long-context reasoning by deferring memory updates to the inference phase, demonstrating notable performance. However, since a full memory update is enforced as a mandatory step in every inference, it inevitably introduces substantial redundant information when processing most simple queries, leading to significant computational overhead and resource waste. Consequently, developing multi-granular memory storage strategies coupled with an adaptive, dynamic scheduling mechanism has become a key challenge to enhancing the efficiency of LLM-based agents. To address this, our proposed hybrid memory framework, HyMem, establishes a two-tier storage architecture consisting of event-level summaries (Level-1 memory) and raw dialogue text (Level-2 memory). By dynamically selecting the appropriate memory granularity based on query complexity, HyMem not only preserves answer accuracy but also significantly improves inference efficiency.
\section{Method}
\begin{figure*}[t] 
  \centering
  \includegraphics[width=\textwidth]{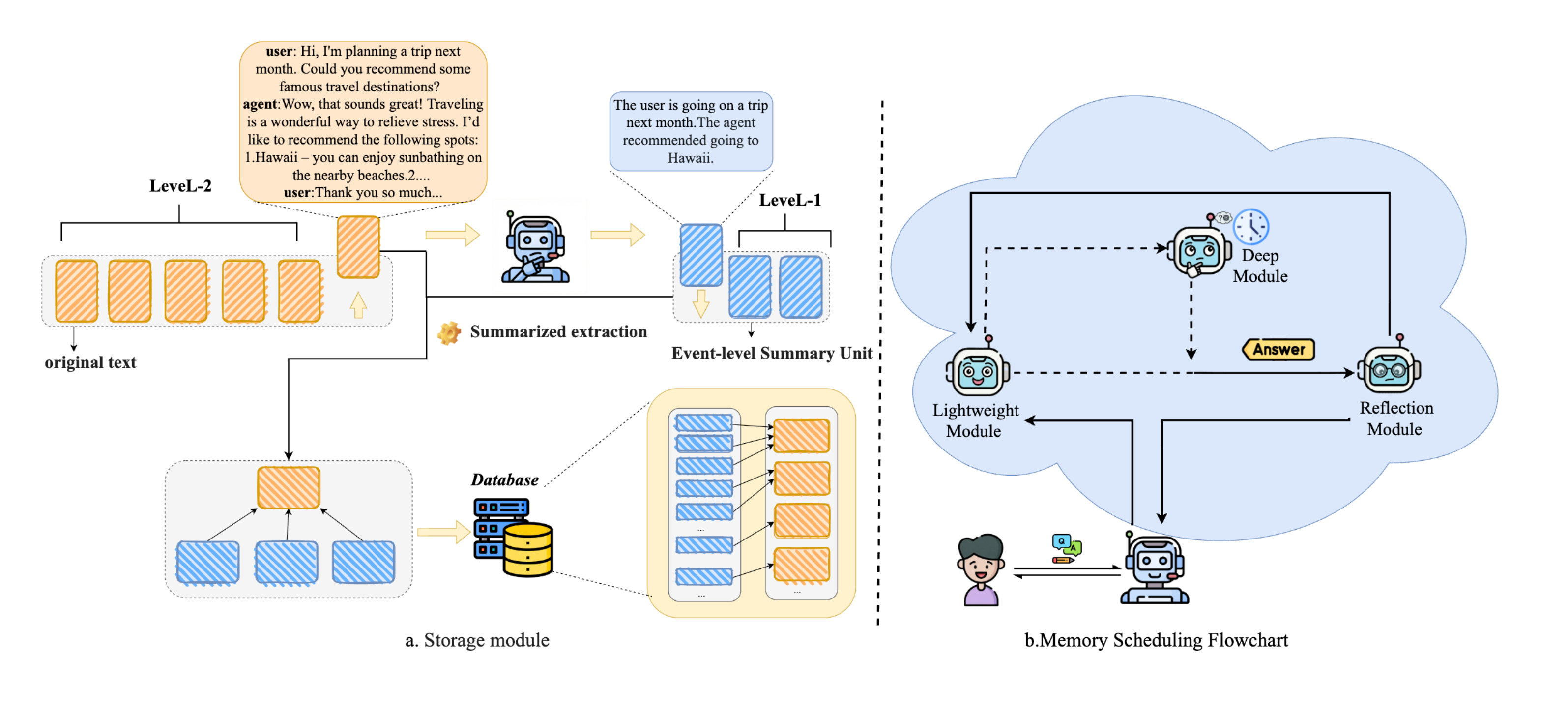}
  \caption{(a) Workflow of the Memory Storage Module. This diagram illustrates the complete pipeline for constructing the dual-layer memory structure: generating core summaries (Level-1 memory) from pre-partitioned event units of raw dialogues (Level-2 memory), establishing many-to-one links to support memory backtracking during the recall stage, before final persistent storage in the database. (b) Inference Workflow within the Memory Recall Module. Queries are first routed to a lightweight module, with selective activation of a deep module when necessary, followed by iterative optimization and final review by the Reflection Module to produce the output.}
  \label{fig:figure2}
\end{figure*}
We propose HyMem, a memory architecture comprising four core components: a memory storage module, a lightweight memory module, a deep memory module, and a reflection module. The overall design aims to efficiently simulate the multi-level memory processing observed in humans, enabling agents to store and retrieve information flexibly and effectively in long-dialogue scenarios.
\subsection{Memory Storage Module}

Current approaches often treat the majority of content in long dialogues as redundant, arguing that such redundancy not only increases token consumption during inference but also obscures critical information. However, information deemed low-value may not be useless, and over-compression inevitably leads to the loss of useful content—an issue we will analyze in detail in Section~\ref{sec:compression_loss}. While we acknowledge the value of distilled memory for efficient storage and question answering—experiments on the LOCOMO benchmark show that in approximately 70\% of cases, answers generated using distilled memory combined with a naive RAG strategy suffice—the remaining 30\% reveal critical bottlenecks. Among these failures, about 15\% and 5\% are attributable to retrieval errors and information loss due to compression, respectively.
As shown in Figure~\ref{fig:figure2}(a), to flexibly accommodate diverse scenario requirements, HyMem’s memory storage module first partitions the raw dialogue into discrete event units based on discussed topics. To maintain coherence between events, we allow content overlap across different units. For each abstracted event, core information is extracted. Specifically, the pipeline distills the general progression of the event and key elements that can be directly associated with queries—such as context, time, location, and participants—into a coherent summary, while omitting dialogue details, causal relationships, and social conventions as redundant content. This forms a Level-1 memory unit $s_{i}$. The corresponding raw dialogue text for each event is preserved as a Level-2 memory unit $p_{i}$.

During the storage phase, the system performs two main operations: First, each Level-1 memory unit is vectorized using an embedding model $E$ , such that $E(s_{i})=e_{i}$, to support efficient retrieval in the lightweight module. Unlike \cite{tan2025prospect}, Level-1 memory units in HyMem serve not only as retrieval entries but also as raw material for context construction, enabling efficient recall and lightweight reasoning throughout the inference process. Second, explicit links are established between each Level-1 memory unit and its corresponding Level-2 memory unit. This design allows the deep memory module, upon detecting a potential logical association between a query and $s_{i}$, to backtrack to the detailed content in Level-2 memory, thereby reconstructing a more informative and reliable context. Both types of memory units are stored synchronously in the memory database, providing multi-granular support for retrieval and reasoning.

\subsection{Light Memory Module}
During the inference process, we maintain a memory pool $M_{i}$, which is initially empty and used to store intermediate answers generated during the question-answering process. The initial user query is denoted as $q_{0}$, and $q_{i}$ represents the intermediate query at the i -th reasoning step; in the first iteration, $q_{i}$is initialized to $q_{0}$.

\begin{figure*}[t] 
  \centering
  \includegraphics[width=\textwidth]{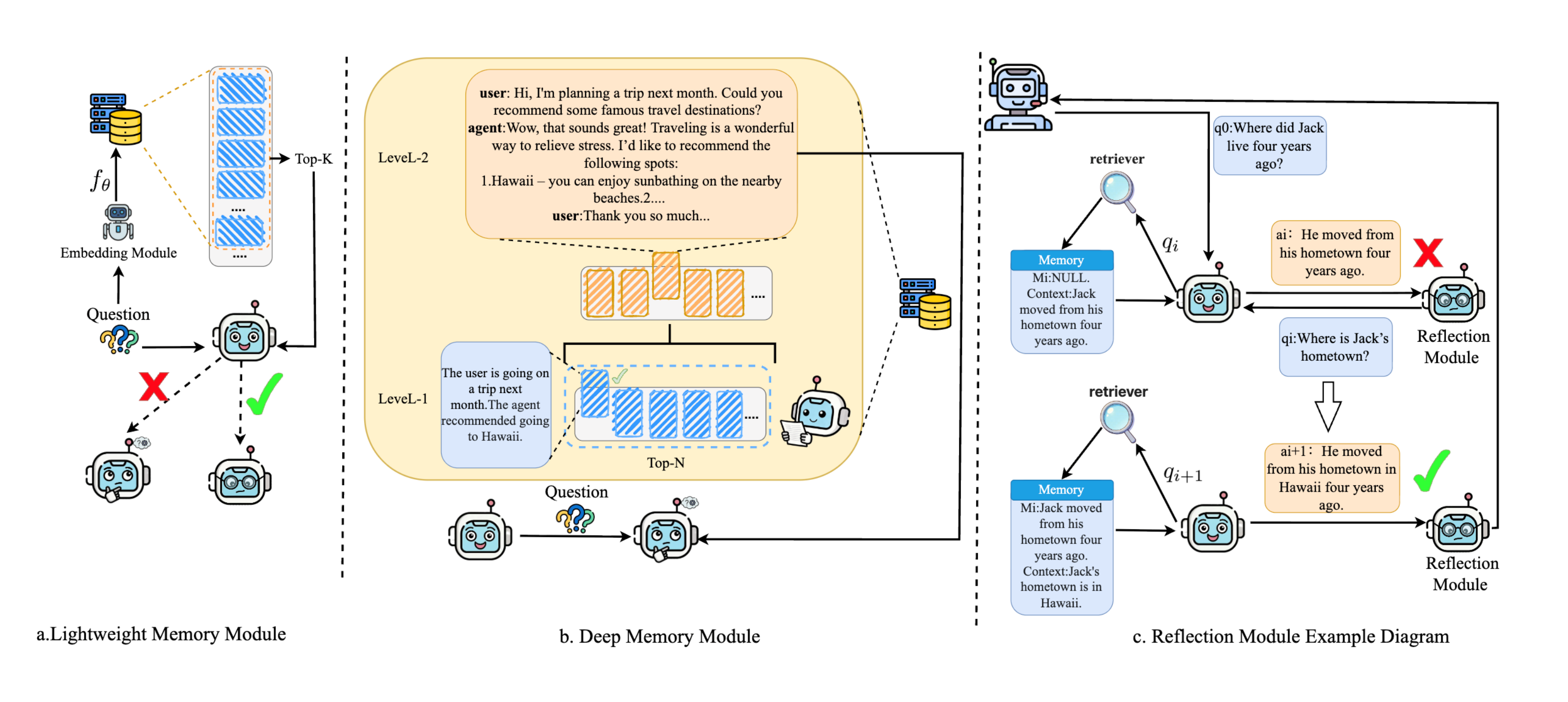}
  \caption{(a) Workflow of the Lightweight Module. (b) Context Reconstruction via the Deep Memory Module by Backtracking from Level-1 to Level-2 Memory. (c) Process of Review and Iterative Optimization by the Reflection Module.}
  \label{fig:figure3}
\end{figure*}

As illustrated in Figure\ref{fig:figure2}(b), we introduce a dynamic memory scheduling mechanism that adaptively responds to input queries. In accordance with the cognitive economy principle—which mimics energy-efficient human recall—the incoming query $q_{i}$ is first sent to the lightweight memory module L for vectorization. As shown in Figure\ref{fig:figure3}(a), module L retrieves the top-k most relevant segments from the Level-1 memory units using cosine similarity, constructs a lightweight context, and attempts to generate an intermediate answer. This process is characterized by low latency and minimal computational overhead, making it suitable for most simple query scenarios.
Additionally, the lightweight memory module evaluates the contextual state to determine whether to invoke the deep memory module. If it detects contextual incompleteness—defined as a "forgetting" state due to either retrieval failure or information loss—the deep memory module is dynamically activated, and $q_{i}$ is forwarded for more precise retrieval and Level-2 memory reconstruction. If the context is deemed complete, the answer is sent directly to the reflection module for further evaluation, and the memory pool $M_{i}$ is updated accordingly.
\subsection{Deep Memory Module}
As illustrated in Figure~\ref{fig:figure3}(b), the deep memory module consists of a retriever $f_\gamma$ and a generator $G_\gamma$. Unlike the lightweight module, which relies solely on semantic similarity for retrieval, this module adopts a two-stage retrieval strategy: first, it performs coarse-grained recall based on semantic similarity to select the top-N candidate events, narrowing the search scope; subsequently, it introduces an LLM-based self-retrieval mechanism to deeply identify key events $s_{k}$ in the Level-1 memory that exhibit semantic relevance, logical connection, or causal dependency with the query $q_{i}$. The retriever then backtracks to the corresponding Level-2 memory unit $p_{k}$ via pre-established links, reconstructing a context with comprehensive information. This context, along with the original query $q_{0}$, is fed into the generator $G_\gamma$ to produce the final answer, and the memory pool $M_{i}$ is updated accordingly.

HyMem defers reasoning about inter-event relationships and query decomposition to the reflection stage. During a single retrieval process, each event is treated as an independent, atomic memory unit. Thanks to the decoupled architecture of the retriever and generator, the retriever can process different batches of Level-1 memory units in parallel, without considering dependencies among units. Specifically, the candidate set obtained from coarse recall can be partitioned by a batch size $d$ :
\begin{equation}
S = \bigcup_{i=1}^{m/d} S_i, \quad S_i = \{ s_{(i-1)d+1},\ s_{(i-1)d+2},\ \ldots,\ s_{id} \}
\end{equation}.
Our approach effectively avoids the $O(N)$ time complexity inherent in traditional streaming memory processing schemes\cite{yu2025memagent}. The overall inference complexity of the system primarily depends on the number of iteration rounds in the reflection phase and the concurrency of LLM calls, thereby significantly improving processing efficiency while maintaining retrieval quality.

\subsection{Reflection Module}
Before the system outputs the final answer, the reflection module R conducts a comprehensive completeness check. Unlike the lightweight contextual completeness assessment performed by the memory module, the reflection module autonomously plans subsequent reasoning steps based on the response completeness. Specifically, this module rigorously evaluates whether the query has been decomposed into minimal answerable units and whether each unit has been correctly addressed. If further responses are possible, the reflection module decomposes the original problem into sub-problems and reformulates the query, generating $q_{i+1}$ for the next round of retrieval. If the problem is fully resolved, the process terminates and the final answer is produced.

This mechanism is particularly critical for complex queries, whose answers may depend on multiple memory segments. As illustrated in Figure~\ref{fig:figure3}(c), although memory units may be independent, the chained combination of queries can give rise to progressive causal or dependency relationships. Since each retrieval round is treated as an atomic operation, a single retrieval may only yield a superficial answer, while underlying dependencies could lead to incomplete responses. The reflection and query rewriting process effectively mitigates this limitation, substantially enhancing the system’s ability to handle complex queries.

\section{Experiment}
Our experiments consist of six parts: (1) as shown in Sections~\ref{subsec:main-result} and~\ref{subsec:additional-datasets}, we conduct detailed comparisons between HyMem and representative baselines on LOCOMO and LongMemEval; (2) in Section~\ref{subsec:efficiency}, we analyze the efficiency--effectiveness trade-off of HyMem in terms of token usage and inference latency; (3) we compare HyMem with Naive RAG under different budgets, considering both performance and computational cost; (4) we perform ablations to quantify the contributions of Level-1/Level-2 granularity, the lightweight/deep memory modules, and the reflection module; (5) we study information loss induced by memory pre-compression and validate our event-level compression strategy; and (6) we sweep the retrieval hyperparameter $k$ in the lightweight memory module.
\subsection{Main Result}\label{subsec:main-result}
On the LOCOMO dataset, we set the retrieval count k of the lightweight module to 10 and the coarse retrieval count N of the deep memory module to 30 for experimentation. As summarized in Table~\ref{tab:table1}, our method demonstrates competitive performance on the single-hop task (ranking second), while consistently exceeding all baseline models across all other task categories as well as in overall average accuracy. Notably, HyMem achieves particularly strong results on multi-hop and open-domain tasks, with accuracies of 88.16\% and 77.08\%, respectively. Even when compared to baseline methods with full context access, it yields substantial improvements of 10.46\% and 5.20\% points, clearly validating the efficacy of the proposed deep memory module in handling complex reasoning challenges. Furthermore, on the LongMemEval-S benchmark, our approach attains a leading score of 78.80\%, surpassing all baseline methods. These results indicate that our method exhibits strong competitive advantages in scenarios involving ultra-long contexts.
\begin{table*}[t]
\centering
\caption{Comparison of LLM-as-a-Judge scores (\%) between HyMem and each baseline on the LOCOMO dataset and LongMemEval-s.}
\label{tab:table1}
\resizebox{\textwidth}{!}{
\begin{tabular}{c|c|cccccc|c}
\toprule
\textbf{} & \textbf{Method} &  & \multicolumn{5}{c|}{\textbf{LOCOMO}}  & \textbf{LongMemEval-s} \\
\midrule
\textbf{} & \textbf{} & \textbf{} Avg Tokens & Single Hop & Multi Hop & Open Domain & Temporal & Overall & Overall \\
\midrule
\multirow{5}{*}{\rotatebox[origin=c]{90}{\textbf{gpt-4o-mini}}}
  
    & A-Mem & 2.7k & 40.22 & 19.01 & 55.10 & 50.12 & 49.01 & 62.60 \\
    & Mem0 & 1.7k & 67.13 & 51.15 & 72.93 & 55.51 & 66.88 & 53.61 \\
    & Zep & 1.4k & 74.11 & 66.04 & 67.71 & 79.76 & 75.14 & 61.20 \\
    & LightMem & 0.8k & 76.61 & 67.02 & 45.83 & 76.32 & 72.99 & 68.64 \\
    & Nemori & 4.7k & 82.10 & 65.30 & 44.80 & 71.00 & 74.4 & 64.20 \\
    & Full-Context & 21.4k & 80.14 & 46.89 & 58.33 & 90.49 & 77.51 & 56.80 \\
    & Our & 1.2k & 72.34 & 80.06 & 71.88 & 88.82 & 82.92 & 75.00 \\
\midrule
\multirow{11}{*}{\rotatebox[origin=c]{90}{\textbf{gpt-4.1-mini}}}
  
    & Zep & 1.4k & 79.43 & 69.16 & 73.96 & 83.33 & 79.09 & 71.12 \\
    & Mem0 & 1.0k & 62.41 & 57.32 & 44.79 & 66.47 & 62.47 & 66.40 \\
    & Mirix & 5k & 85.11 & 83.70 & 65.62 & 88.39 & 85.38 & 56.80 \\
    & Nemori & 4.4k & 84.90 & 75.10 & 51.00 & 77.60 & 79.40 & 74.60 \\
    & MemU & 4.0k & 74.91 & 72.34 & 43.61 & 54.17 & 66.67 & 38.40 \\
    & Naive RAG(k=3)  & 2.5k & 70.92 & 69.47 & 65.62 & 75.74 & 72.97 & 67.20 \\
    & Naive RAG(k=5)  & 4.1k & 72.69 & 74.14 & 68.75 & 81.21 & 77.40 & 72.80 \\
    & Full-Context & 21.4k & \textbf{88.53} & 77.70 & 71.88 & 92.70 & 87.52 & 58.20 \\
    & Our & 1.5k & 85.15 & \textbf{88.16} & \textbf{77.08} & \textbf{92.98} & \textbf{89.55} & \textbf{78.80} \\
\bottomrule
\end{tabular}
}
\end{table*}

\subsection{Efficiency Analysis}\label{subsec:efficiency}
On the LoCoMo dataset with the gpt-4o-mini backbone, we compare HyMem with other mainstream baselines in terms of average per-sample inference latency and average token consumption per question. As shown in Table ~\ref{tab:locomo-efficiency-score}, HyMem achieves the best score and significantly outperforms other methods in average per-sample inference efficiency; relative to Mem0 and SimpleMem, HyMem reduces retrieval-stage latency by 83.3\% and 74.9\%, respectively. We attribute these gains to HyMem’s dynamic scheduling mechanism, which further decreases inference latency. In most query scenarios, the system relies only on lightweight modules for memory recall and context construction, and selectively activates the deep-memory module only for complex questions, thereby avoiding the high average latency induced by always using complex retrieval strategies. These results validate the effectiveness of HyMem’s dynamic memory scheduling strategy. By adaptively allocating computational resources according to query complexity, HyMem substantially reduces average computation cost while achieving an optimal balance between performance and efficiency.
\begin{table}[H]
  \centering
  \caption{Inference latency and token consumption comparison on the LOCOMO dataset.}
  \small
  \setlength{\tabcolsep}{5pt}
  \renewcommand{\arraystretch}{1.2}

  \resizebox{\linewidth}{!}{%
  \begin{tabular}{l|cccccccc}
    \toprule
    \textbf{Metric} 
      & \textbf{A-Mem} & \textbf{LightMem} & \textbf{Mem0} & \textbf{SimpleMem} & \textbf{REMem} & \textbf{RF-mem} & \textbf{Mem1} & \textbf{HyMem} \\
    \midrule
    \textbf{Inference Time (s)} 
      & 796.7 & 577.1 & 583.4 & 388.3 & 312.0 & 497.0 & 283.9 & \textbf{97.4} \\
    \textbf{Cost Tokens (k)} 
      & 2.7 & 0.8 & 1.0 & 5.4 & 9.4 & 9.4 & 9.6 & 1.2 \\
    \textbf{LLM Score} 
      & 49.01 & 72.99 & 66.88 & 71.43 & 74.02 & 76.49 & 72.60 & \textbf{82.92} \\
    \bottomrule
  \end{tabular}%
  }

  \label{tab:locomo-efficiency-score}
\end{table}

\subsection{Comparison with a Naive RAG}\label{subsec:naive-rag}
\begin{figure*}[t] 
    \centering
    \begin{subfigure}{0.24\textwidth}
        \centering
        \includegraphics[width=\linewidth]{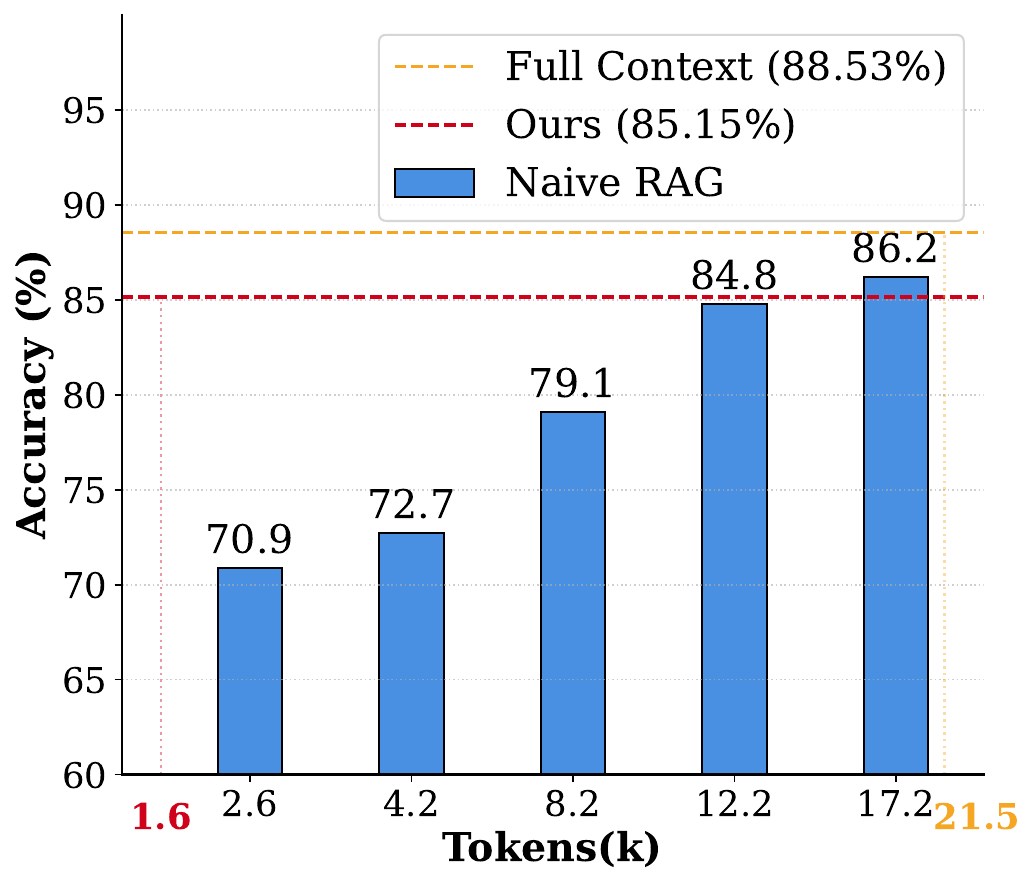}
        \caption{Single Hop tasks.}
        \label{fig:rag1}
    \end{subfigure}
    \hfill 
    \begin{subfigure}{0.24\textwidth}
        \centering
        \includegraphics[width=\linewidth]{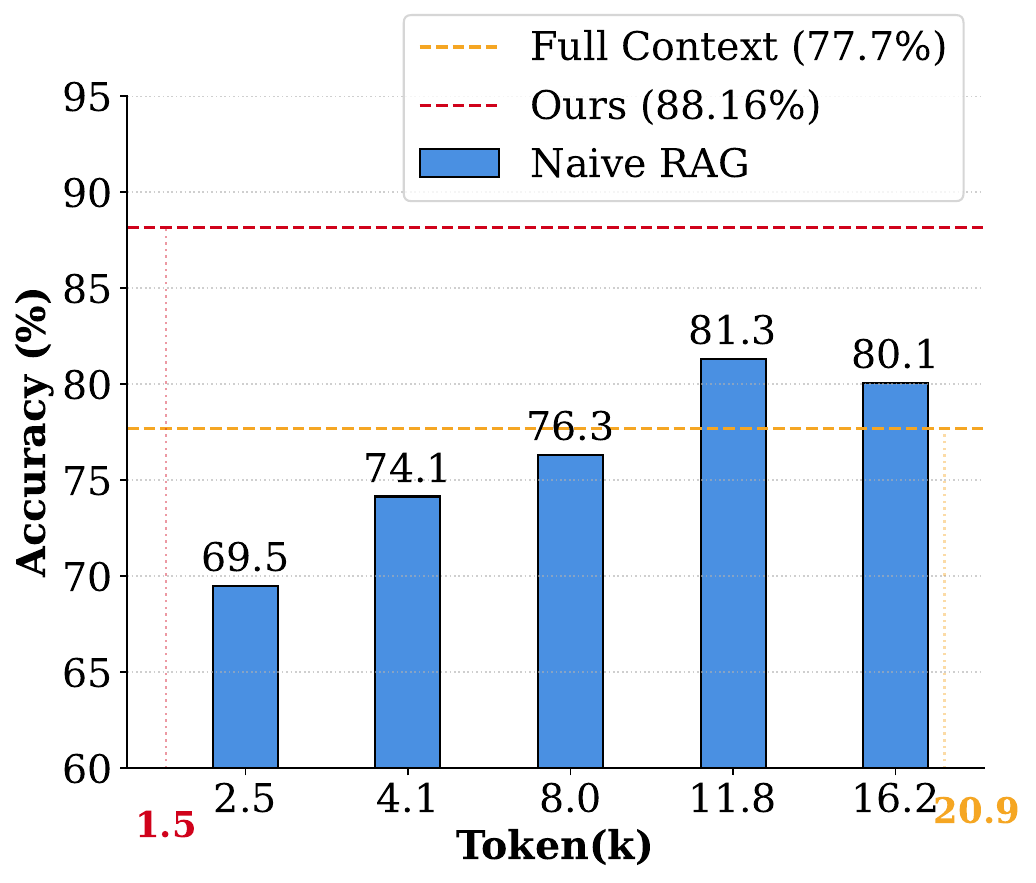}
        \caption{Multi Hop tasks.}
        \label{fig:rag2}
    \end{subfigure}
    \hfill
    \begin{subfigure}{0.24\textwidth}
        \centering
        \includegraphics[width=\linewidth]{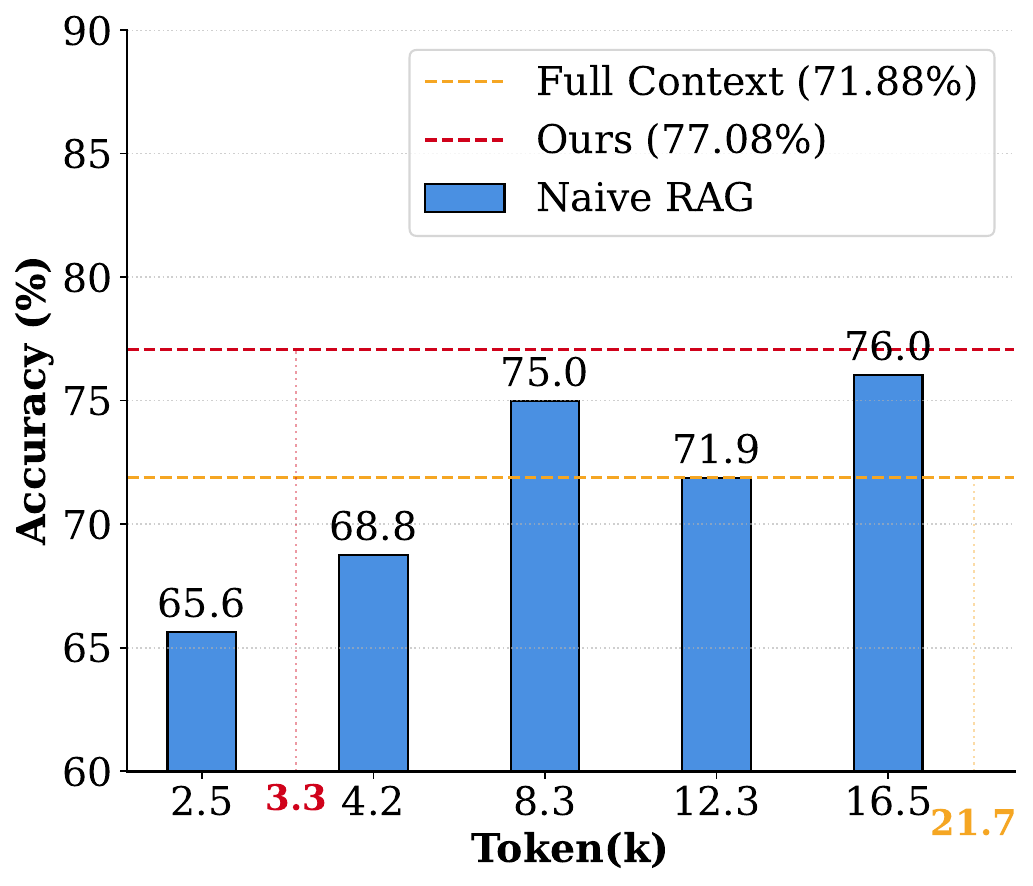}
        \caption{Open Domain tasks.}
        \label{fig:rag3}
    \end{subfigure}
    \hfill
    \begin{subfigure}{0.24\textwidth}
        \centering
        \includegraphics[width=\linewidth]{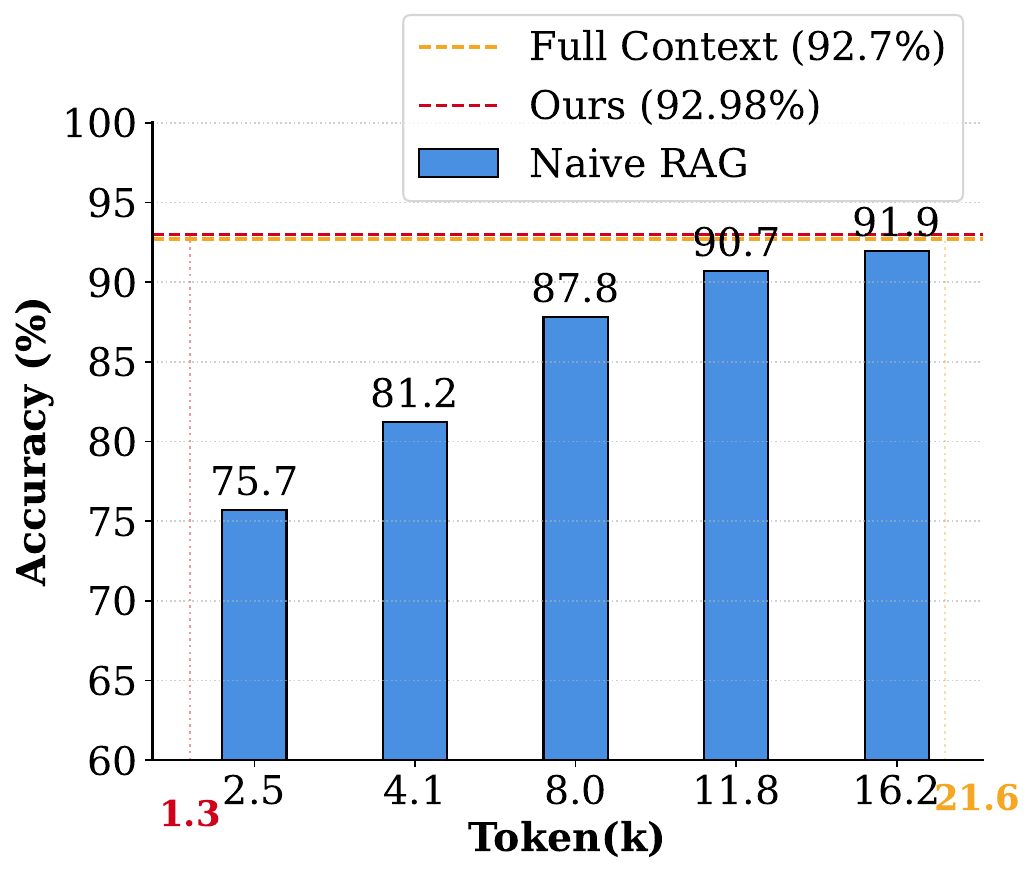}
        \caption{Temporal tasks.}
        \label{fig:rag4}
    \end{subfigure}
    
    \caption{Comparison of performance and average token usage on the four LOCOMO task categories for Naive RAG with different retrieval $k$ values, Full Context, and our method.}
    \label{fig:overall_rag}
\end{figure*}

We compare the efficiency and effectiveness of Naive RAG approaches with our proposed method under varying numbers of retrieved items $k$ . The experiments employ text-embedding-3-small as the embedding model and are evaluated on four types of tasks from the LOCOMO dataset. For Naive RAG, we test accuracy and inference token consumption for $k \in \{3,\, 5,\, 10,\, 15,\, 20\}$. To maximize baseline performance, retrieval is performed directly on the original memory segments rather than compressed representations. Our method uses the same embedding model but fixes $k=10$ in the retrieval module.
As shown in Figure~\ref{fig:overall_rag}, the accuracy of Naive RAG initially increases with larger $k$ , which is expected since a higher recall count improves the likelihood of retrieving relevant events. However, beyond a certain threshold, performance plateaus or even declines. This observation aligns with findings from \cite{fang2025lightmem,liu2024lost,pan2025secom}, and our experiments confirm that excessively redundant context leads to diminishing returns. In contrast, HyMem achieves significantly lower inference token overhead while attaining higher accuracy than Naive RAG under equivalent cost settings—even outperforming the full-context baseline.
\label{sec:main_result}

\subsection{Ablation Studies}

We conduct a comprehensive ablation study on the LOCOMO dataset to evaluate the impact of memory granularity in HyMem's storage module—core summaries as Level-1 memory versus raw text as Level-2 memory—along with the three inference-time modules: the lightweight memory module, deep memory module, and reflection module. Table~\ref{tab:ablation-memory-modules} summarizes the experimental results.
Notably, removing both the deep memory module and Level-2 memory considerably reduces inference token usage (by 700–800 tokens), but at the cost of a 17.40\% drop in accuracy, which we attribute to the coarsening of information granularity and suboptimal retrieval performance. In contrast, removing the lightweight memory module yields a slight accuracy improvement of 2.20\%; however, this forces all queries to be processed exclusively by the deep memory module, increasing the average token consumption to 3192. Since approximately 70\% of queries can be handled efficiently by the lightweight module in practice, omitting this component leads to substantial computational inefficiency.
Furthermore, ablating the reflection module results in accuracy reductions of 2.21\%, 4.17\%, 0.23\%, and 1.09\% under different configurations, highlighting its crucial role in enhancing the agent's ability to manage complex multi-turn dialogues—a process that also incurs a modest rise in token usage. Lastly, utilizing Level-2 memory in place of Level-1 memory improves accuracy by about 6.34\%, albeit at the expense of increasing average token consumption by roughly fivefold.
\begin{wraptable}{r}{0.5\textwidth}
  \centering
  \small
  \setlength{\tabcolsep}{3pt}
  \renewcommand{\arraystretch}{1.2}
  \begin{tabular}{
        >{\centering\arraybackslash}m{0.8cm}
        >{\centering\arraybackslash}m{0.8cm}|
        >{\centering\arraybackslash}m{0.45cm}
        >{\centering\arraybackslash}m{0.45cm}
        >{\centering\arraybackslash}m{0.45cm}|
        >{\centering\arraybackslash}m{1.2cm}
        >{\centering\arraybackslash}m{1.2cm}
    }
        \toprule
        \multicolumn{2}{c|}{\textbf{Storage}} &
        \multicolumn{3}{c|}{\textbf{Retrieval}} &
        \multirow{2}{*}{\textbf{Overall}} &
        \multirow{2}{*}{\textbf{Token}} \\
        Level-1 & Level-2 & L & D & R & & \\
        \midrule
        \checkmark & \checkmark & \checkmark & \checkmark & \checkmark & 89.55 & 1540 \\
        \checkmark &  & \checkmark & & \checkmark & 72.15 & 850 \\
        \checkmark & \checkmark & & \checkmark & \checkmark & 91.75 & 3192 \\
        \checkmark & \checkmark & \checkmark & \checkmark & & 87.34 & 1411 \\  
        \checkmark &  & \checkmark & & & 67.98 & 794 \\
        \checkmark & \checkmark & & \checkmark & & 91.16 & 2947 \\
        & \checkmark & \checkmark & & \checkmark & 78.49 & 4125 \\
        & \checkmark & \checkmark & & & 77.40 & 4046 \\
        \bottomrule
  \end{tabular}

  \caption{Impact of Memory Granularity (Core Summary Level-1, Original Text Level-2) and Retrieval Modules (Lightweight, Deep, Reflection) on Overall (\%) and Avg Token Consumption.}
  \label{tab:ablation-memory-modules}
\end{wraptable}
\vspace{-10pt}
\subsection{Compression Loss Analysis}
\label{sec:compression_loss}
\begin{figure*}[t] 
    \centering
    \begin{subfigure}{0.24\textwidth}
        \centering
        \includegraphics[width=\linewidth]{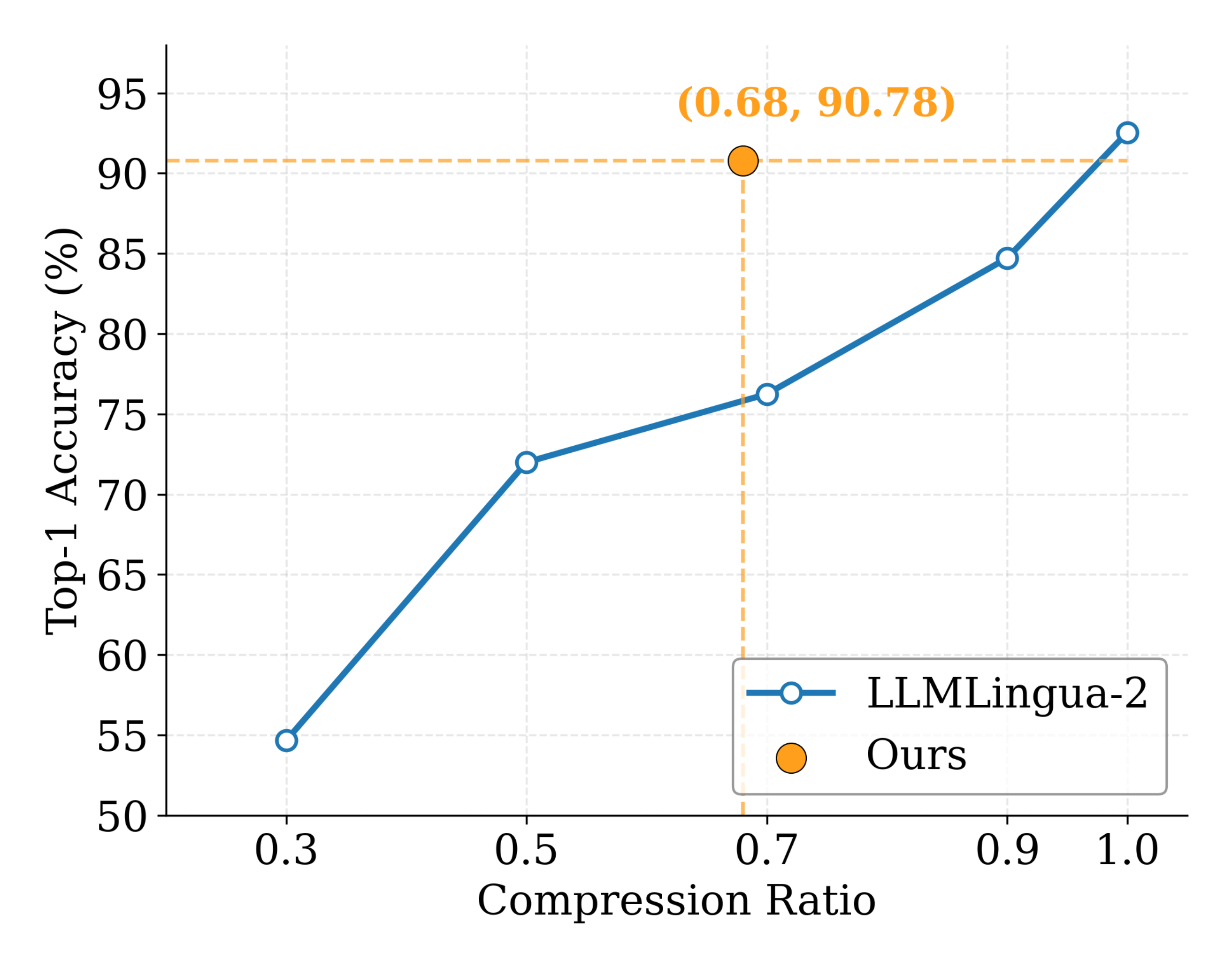}
        \caption{Single Hop tasks.}
        \label{fig:error1}
    \end{subfigure}
    \hfill 
    \begin{subfigure}{0.24\textwidth}
        \centering
        \includegraphics[width=\linewidth]{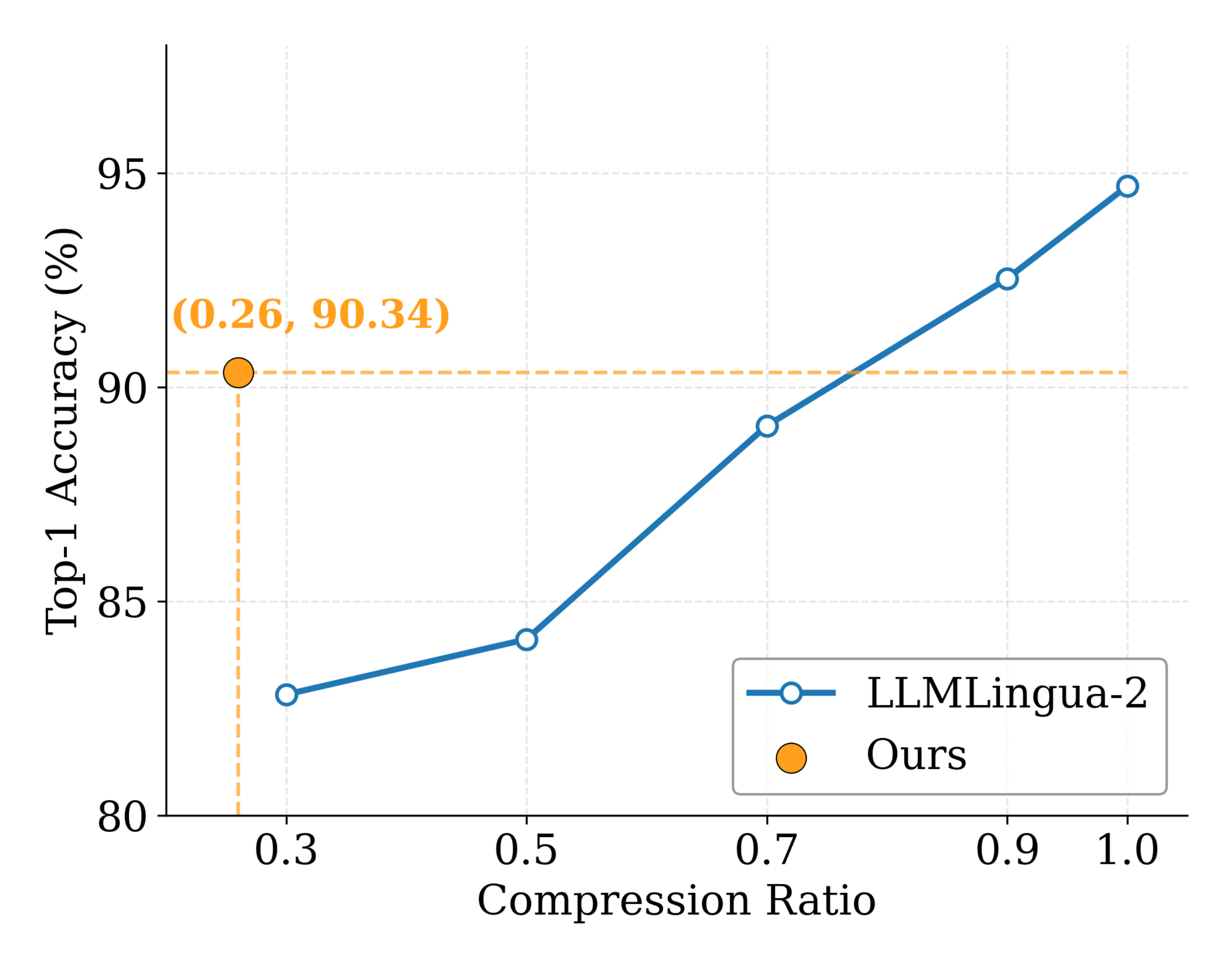}
        \caption{Multi Hop tasks.}
        \label{fig:error2}
    \end{subfigure}
    \hfill
    \begin{subfigure}{0.24\textwidth}
        \centering
        \includegraphics[width=\linewidth]{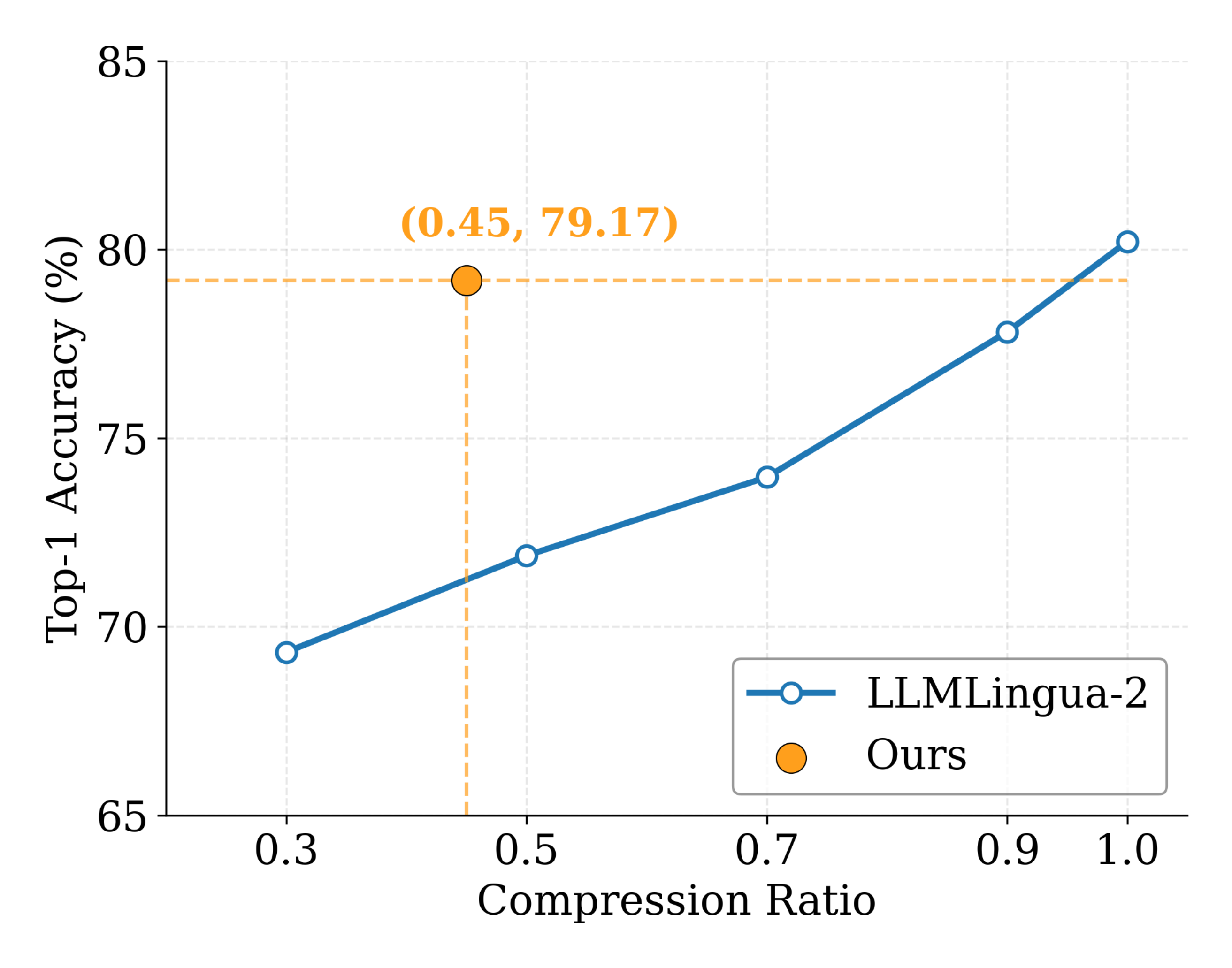}
        \caption{Open Domain tasks.}
        \label{fig:error3}
    \end{subfigure}
    \hfill
    \begin{subfigure}{0.24\textwidth}
        \centering
        \includegraphics[width=\linewidth]{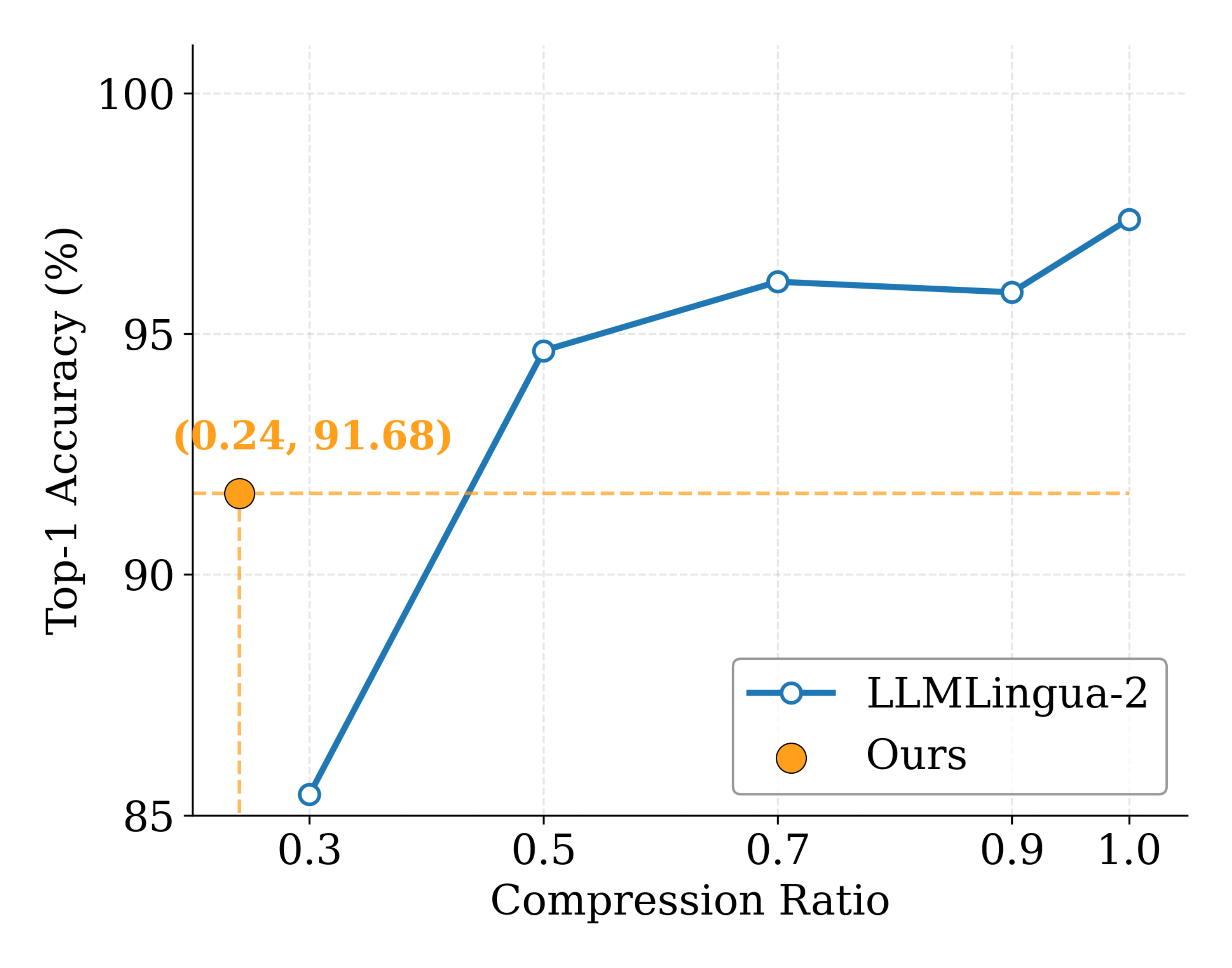}
        \caption{Temporal tasks.}
        \label{fig:error4}
    \end{subfigure}

    \caption{Evaluating accuracy loss under different compression ratios using the LLMLingua-2 pre-compression method.}
    \label{fig:overall_error}
\end{figure*}

We hypothesize that the errors made by agents in long-context tasks mainly stem from the following factors: (1) Retrieval Errors: Retrieved passages are irrelevant to the query or incomplete in content; (2) Over-compression: Information loss caused by aggressive compression during memory storage, leading to insufficient evidence retrieved for reasoning; (3) Model Reasoning Errors: The model still outputs incorrect answers even when provided with the complete context. This is often due to excessive textual redundancy, which hinders the formation of effective context associations, or due to the inherent complexity of the query exceeding the model's reasoning capabilities; (4) Dataset Errors: The dataset itself contains intrinsic flaws.
Since the last two types of errors (Model and Dataset Errors) are inherent to the model and dataset, we isolate and evaluate the impact of compression granularity solely by removing the retrieval module and directly providing the compressed 'ground-truth' passages as context to the model. Specifically, we employ the LLMLingua-2 model to pre-compress the original text and evaluate the performance across the four task types in the LOCOMO benchmark under varying compression ratios, $ratio \in \{0.3, 0.5, 0.7, 0.9, 1.0\}$. The setting with a compression ratio of 1.0 (the full, uncompressed passage is provided) serves as an approximate upper bound of the model's performance on LOCOMO.
As shown in Figure~\ref{fig:overall_error}, after eliminating noise introduced by retrieval, task accuracy declines as the compression ratio decreases, validating our initial hypothesis that fine-grained memory is critical for certain QA tasks. Further experiments demonstrate that our proposed event-level compression method, which reduces memory length to approximately 30\% of the original text, significantly outperforms LLMLingua-2 in accuracy—even under the same or more aggressive compression ratios. This indicates that our compression strategy more effectively preserves information crucial for downstream reasoning while efficiently controlling textual redundancy.
\subsection{Hyperparameter Analysis}
As shown in Table~\ref{tab:hyperparameter-table-horizontal}, initially, as the recall number $k$ in the lightweight memory module increases, model accuracy improves only modestly and does not exhibit significant fluctuations (the variance in accuracy from $k=1$ to $k=30$ is approximately 0.60). This behavior can be attributed to HyMem's dynamic retrieval mechanism: when the retrieval results from the lightweight memory module fail to meet the query requirements, the deep memory module is effectively activated to perform supplementary retrieval, thereby ensuring overall response capability. However, higher values of $k$ improve the recall success rate of the lightweight module, significantly reducing reliance on the deep memory module and causing its average invocation rate to decline. Given that the primary token consumption during inference stems from the LLM-based autonomous retrieval in the deep memory module, the reduced invocation frequency leads to a decline in overall token usage when $k<10$ . Notably, once $k$ exceeds a certain threshold ($k>10$ ), the invocation rate of the deep memory module tends to saturate. Beyond this point, further increasing $k$ yields diminishing returns in terms of performance gain, while introducing additional token overhead due to the expanded retrieval content of the lightweight module.
\vspace{-5pt}
\begin{table}[H]
  \centering
    \caption{Effect of recall hyperparameter $k$ on Overall (\%), Token Consumption, and Deep Memory Module Average Invocation Rate.}
  \small
  \setlength{\tabcolsep}{5pt}
  \renewcommand{\arraystretch}{1.2}

  \begin{tabular}{c|cccccccc}
    \toprule
    \textbf{k} 
      & 1 & 3 & 5 & 7 & 10 & 12 & 15 & 20 \\
    \midrule
    \textbf{Overall} 
      & 87.40 & 88.25 & 88.63 & 88.89 & 89.55 & 89.63 & 89.57 & 89.74 \\
    \textbf{Token} 
      & 1966 & 1747 & 1561 & 1558 & 1540 & 1594 & 1647 & 1729 \\
    \textbf{Deep ratio} 
      & 0.43 & 0.32 & 0.29 & 0.27 & 0.22 & 0.22 & 0.22 & 0.21 \\
    \bottomrule
  \end{tabular}

  \label{tab:hyperparameter-table-horizontal}
\end{table}

\section{Conclusion}
We propose HyMem, a hybrid memory framework for large language model agents inspired by the principle of cognitive economy, designed to address the efficiency-performance trade-off in long dialogue scenarios. Unlike traditional static memory approaches, HyMem employs a dual-granularity memory storage and dynamic scheduling mechanism that strategically allocates retrieval and reasoning resources between a lightweight memory module and an activatable deep memory module based on query complexity. Augmented by a reflection module for multi-turn iterative refinement, our framework significantly enhances the flexibility and robustness of memory management. Experimental results show that HyMem achieves state-of-the-art performance across multiple widely used benchmarks. This work underscores the importance of adaptive, resource-aware memory systems in enabling long-term, robust, and efficient agent interactions, and lays a foundation for theoretical and practical advances in scalable dialogue agents.

{
    \small
    \bibliographystyle{unsrtnat}
    \bibliography{neurips_2025}
}

\clearpage
\newpage
\section*{Appendix}
\setcounter{page}{1}
\appendix

\section{More Results}

\subsection{Evaluation on Additional Datasets}\label{subsec:additional-datasets}

\begin{figure}[H] 
  \centering
  \includegraphics[width=\textwidth]{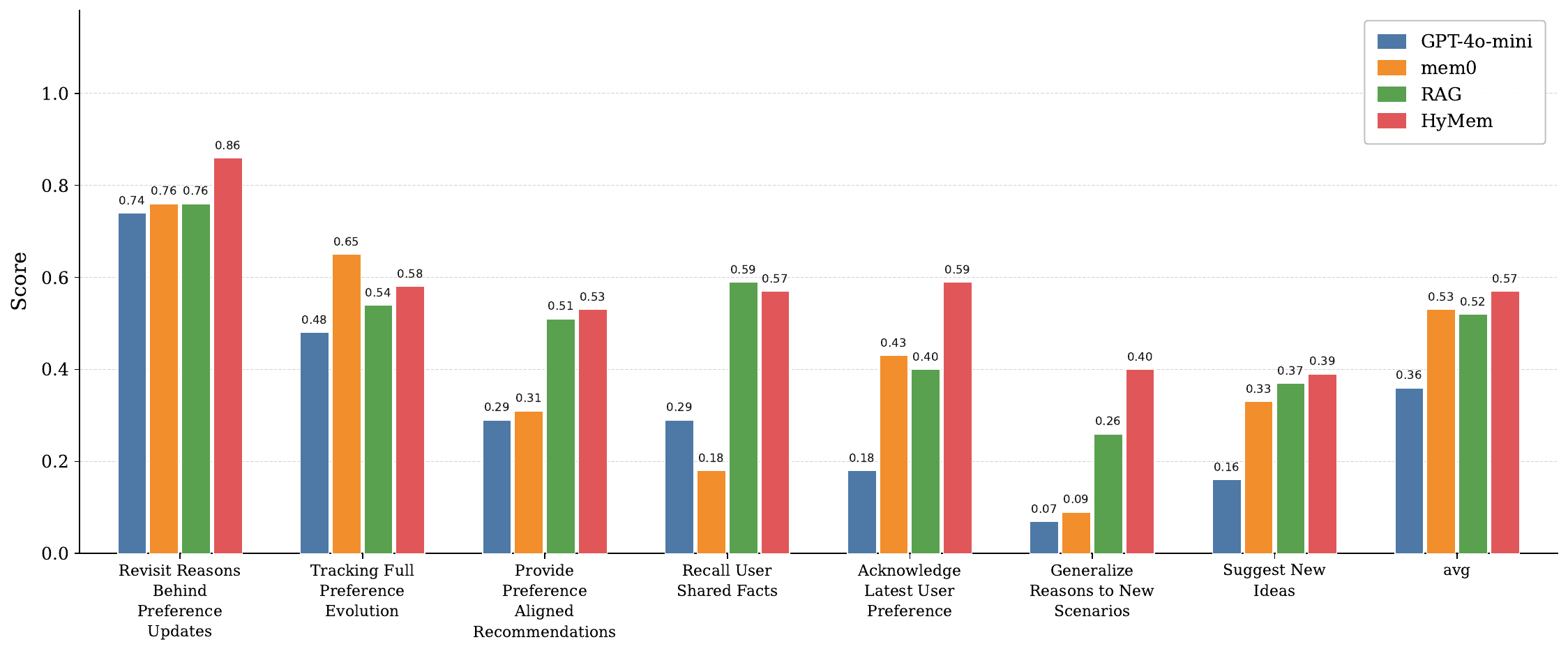}
  \caption{Comparison of performance between mainstream baselines and our method across seven PersonMem-128k task categories.}
  \label{fig:PersonMem}
\end{figure}

\begin{figure}[H] 
  \centering
  \includegraphics[width=\textwidth]{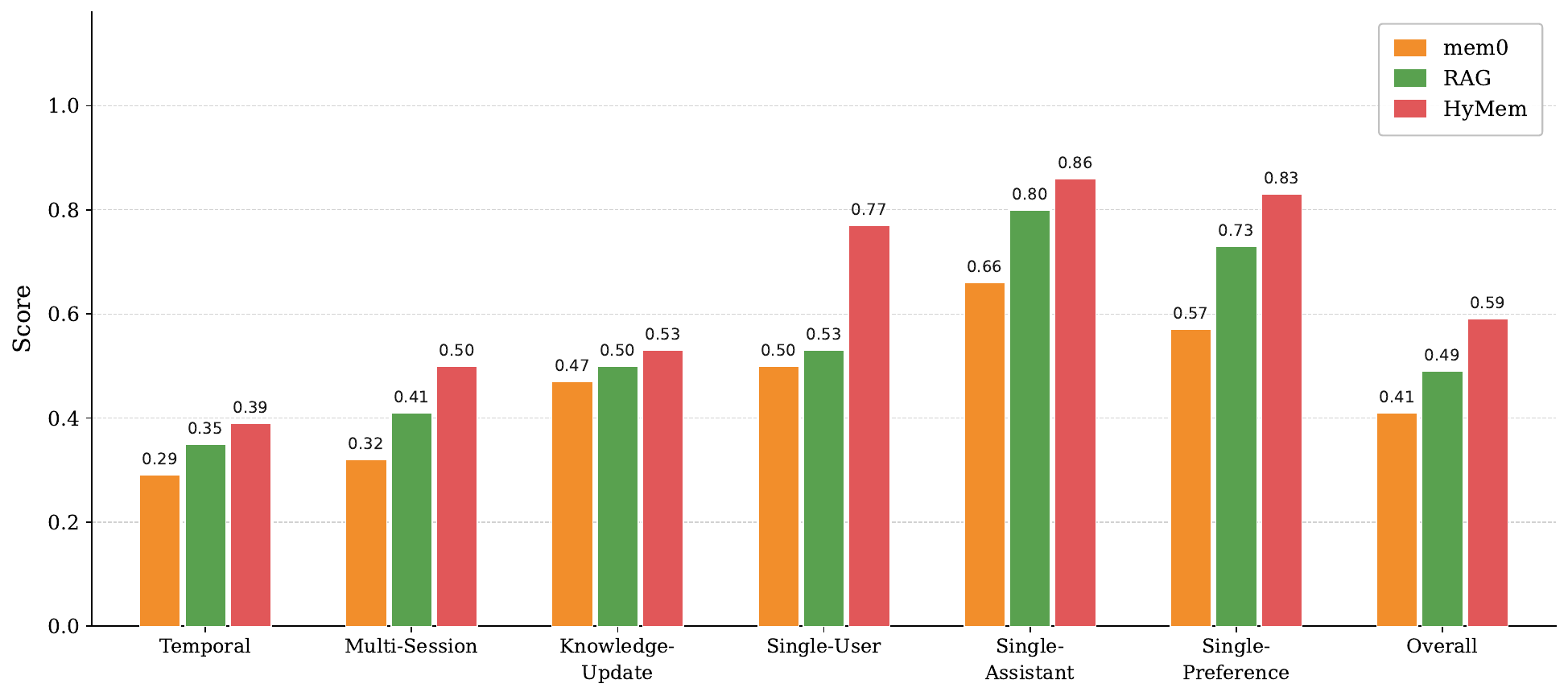}
  \caption{Comparison of performance between mainstream baselines and our method across six LongMemEval-M task categories.}
  \label{fig:LongMemEval-M}
\end{figure}

We further evaluate on the more challenging PersonMem-128k and LongMemEval-M benchmarks. As shown in Figure~\ref{fig:PersonMem} and Figure~\ref{fig:LongMemEval-M}, HyMem outperforms all baselines, further demonstrating its strong applicability even under ultra-long context settings.

\subsection{Comparison of the Lightweight and Deep Modules}
We ablate the reflection module and separately evaluate the impact of the lightweight module and the deep module on both efficiency and effectiveness, and further compare them with the hybrid setting.

As shown in Table ~\ref{tab:light-deep-hybrid-k}, under the same recall budget, complex retrieval improves accuracy by 18.96\% over simple retrieval, indicating that complex retrieval serves as a reliable fallback when simple retrieval fails. However, it introduces an additional 40–90 seconds of latency, due to extra token consumption when processing coarse retrieval outputs. As the coarse retrieval scope expands, accuracy increases slightly, but at the cost of higher token usage and latency. Overall, the hybrid mode (HyMem) attains accuracy close to the best performance of the deep module, while using only one-third of its token budget and reducing latency by 36\%, which further validates the efficiency advantage of dynamic scheduling.

\begin{table}[tb]
  \centering
    \caption{Effect of retrieval budget ($k$) on performance and efficiency for Light, Deep, and Hybrid settings. Total Tokens = Retrieve Tokens + Answer Tokens.}
  \small
  \setlength{\tabcolsep}{5pt}
  \renewcommand{\arraystretch}{1.15}

  \begin{tabular}{c|c|c|cccc}
    \toprule
    \textbf{Module} & \textbf{Retrieve $k$} & \textbf{LLM Score} & \textbf{Retrieve Tokens} & \textbf{Answer Tokens} & \textbf{Total Tokens} & \textbf{Time (s)} \\
    \midrule
    \multirow{1}{*}{Light} 
      & 10 & 67.98 & 0   & 794  & 794  & 76.8 \\
    \midrule
    \multirow{4}{*}{Deep} 
      & 10 & 86.94 & 596  & 1583 & 2179 & 116.8 \\
      & 20 & 89.68 & 941  & 1633 & 2574 & 123.8 \\
      & 30 & 91.16 & 1286 & 1761 & 3047 & 147.8 \\
      & 50 & 91.62 & 2094 & 1973 & 4067 & 170.3 \\
    \midrule
    Hybrid (Light, Deep) 
      & (10, 30) & 87.34 & 179 & 860 & 1039 & 94.6 \\
    \bottomrule
  \end{tabular}

  \label{tab:light-deep-hybrid-k}
\end{table}

\subsection{Error Analysis}
As shown in Table~\ref{tab:error}, we run each experiment three times on the LOCOMO dataset using the gpt-4.1-mini setup, and report the mean and standard deviation to quantify run-to-run stability.

\begin{table}[H]
\centering
\caption{Error analysis on LOCOMO.}
\label{tab:error}
\begin{tabular}{ccccc}
\toprule
Single-hop & Multi-hop & Open-domain & Temporal & Overall \\
\midrule
85.15 & 88.16 & 77.08 & 92.98 & 89.55 \\
86.17 & 85.04 & 78.12 & 91.79 & 88.50 \\
84.04 & 85.35 & 76.04 & 91.43 & 87.86 \\
\midrule
$85.12 \pm 1.07$ & $86.18 \pm 1.72$ & $77.08 \pm 1.04$ & $92.07 \pm 0.81$ & $88.64 \pm 0.85$ \\
\bottomrule
\end{tabular}
\end{table}

\section{Dataset}
\subsection{LOCOMO}
The LoCoMo dataset\cite{maharana2024evaluating} comprises dialogues with an average total length of 9,000 tokens, where each dialogue contains approximately 600 turns and 26,000 tokens. Each dialogue includes around 200 questions, which are categorized into the following types: (1) single-hop questions that can be answered using information from a single turn; (2) multi-hop questions requiring information integration across different turns; (3) temporal reasoning questions designed to assess a model’s ability to reason about sequential events; and (4) open-domain knowledge questions that require combining dialogue context with external knowledge. LoCoMo is particularly suited for evaluating a model’s capacity to handle long-range dependencies and maintain consistency throughout extended interactions.
\subsection{LongMemEval}
The LongMemEval dataset\cite{wu2024longmemeval} focuses on assessing long-context memory capabilities. It includes high-difficulty tasks such as long-text comprehension, multi-stage reasoning, and information integration across segments. The dataset emphasizes model performance in handling ultra-long contexts, complex causal chains, and long-term information retention. Through diverse task settings, LongMemEval enables a more comprehensive evaluation of memory systems with respect to long-range dependencies, detailed tracking, and information retrieval accuracy and robustness, thus providing more stringent standards for practical model application.

\subsection{PersonMem}
The PersonMem dataset \cite{jiang2025know} focuses on evaluating the personalized response capability of LLMs in long-term user-LLM interactions. It contains over 180 simulated user-LLM interaction histories, each comprising up to 60 multi-turn sessions with a total context length of up to 1 million tokens. Each session consists of 15–30 conversation turns across 15 real-world tasks that require personalization, such as travel planning, therapy consultation, and food recommendation. The dataset assesses a model's ability to (1) internalize user traits and preferences from interaction history, (2) track the dynamic evolution of user profiles over time, and (3) generate responses that align with the user's current state in new scenarios. It includes seven types of in-situ user queries (e.g., recalling user-shared facts, acknowledging latest preferences, tracking full preference evolution, and generalizing to new tasks). 

\section{Baselines}
\textbf{A-Mem}\cite{xu2025mem}: A memory system that constructs Zettelkasten-style knowledge graphs from user–agent interactions, dynamically associating notes through embedding similarity analysis and LLM-based reasoning.

\textbf{Zep}\cite{rasmussen2501zep}: A commercial memory API that builds temporal knowledge graphs (Graphiti) from user dialogues and metadata, designed for efficient semantic querying.

\textbf{Mem0}\cite{chhikara2025mem0}: An open-source memory system that incrementally compresses and stores memory facts using LLM-based summarization, with an optional graph memory extension.

\textbf{REMem}\cite{shu2026remem}:A two-phase episodic memory framework that explicitly models experiences as time-aware events. It indexes interactions into a hybrid memory graph that links timestamped natural-language gists with time-scoped factual triples, then performs iterative, tool-augmented “agentic inference” (retrieval + graph exploration) to support episodic recollection and multi-step temporal reasoning. 

\textbf{RF-Mem}\cite{zhang2026evoking}:
An uncertainty-guided dual-path memory retriever for personalized LLMs inspired by the cognitive dual-process theory of \emph{familiarity} vs.\ \emph{recollection}. It performs a probe retrieval and uses the mean similarity score and entropy as a familiarity signal to adaptively switch between (i) fast one-shot top-$K$ retrieval when confidence is high, and (ii) a multi-round recollection path that clusters candidate memories and iteratively expands evidence via query--centroid $\alpha$-mixing in embedding space to reconstruct relevant context. 

\textbf{LightMem}\cite{fang2025lightmem}: A lightweight memory system inspired by human memory models, which employs a three-stage mechanism—sensory memory filtering, topic-aware short-term memory organization, and long-term memory updated during sleep time—to significantly improve memory processing efficiency while maintaining performance.

\textbf{Nemori}\cite{nan2025nemori}: A self-organizing memory architecture inspired by event segmentation theory and the free-energy principle. It autonomously structures conversational streams into semantically coherent episodes via a Two-Step Alignment principle, and enables agents to proactively learn from prediction errors through a Predict–Calibrate principle, facilitating adaptive knowledge evolution.

\textbf{MIRIX}\cite{wang2025mirix}: A modular, multi-agent memory system that integrates six specialized components—Core, Episodic, Semantic, Procedural, Resource Memory, and Knowledge Vault—coordinated via a multi-agent framework for dynamic memory updates and retrieval. It is particularly adept at handling large-scale multimodal inputs to enable deep personalization.

\textbf{Full Context}: Provide the LLM with all the context corresponding to the question.

\textbf{Naive RAG}: Vectorize the content using the qwen3-0.6b-embedding model and store the entire context as memory for subsequent responses.

\section{Limitations}

HyMem achieves encouraging results across a broad range of widely used benchmarks, striking a strong balance between effectiveness and efficiency. Nevertheless, several directions remain for future investigation. First, while HyMem can be readily integrated with both open-source and proprietary foundation models, its ability to plan retrospection paths and identify key events is inherently constrained by the underlying language model’s capabilities; consequently, different LLM backbones may yield variations in generation quality. Second, HyMem currently focuses on text-based interactions. Extending the framework to multimodal inputs (e.g., images and audio) is a promising avenue for future work, as it may enable richer context representations.

\section{Algorithm}
\begin{algorithm}[H]
  \caption{HyMem Memory Retrieval and Reasoning}
  \label{alg:HyMem}
  \begin{algorithmic}[1]
    \STATE \textbf{Input:} Initial query $q_0$, maximum number of iterations $T$
    \STATE \textbf{Initialize:} Memory pool $M_0 \leftarrow$ NULL, iteration counter $i \leftarrow 0$
    \STATE \textbf{Modules:} Light Memory Module $\mathcal{L}=\{$Embedding $E$, Retriever $f_\theta$, Generator $G_\theta\}$, \\ Deep Memory Module $\mathcal{D}=\{$ Retriever $f_\theta$, Retriever $f_\gamma$, Generator $G_\gamma\}$, Reflection module $R$
    \WHILE{not \texttt{done} and $i < T$}
      \STATE $e_{q_i} \leftarrow E(q_i)$ \hfill // Embed query
      \STATE $[s]_k \leftarrow$ Retrieve top-$k$ Level-1 memories using $f_\theta(e_{q_i})$
      \STATE $answer_i, status \leftarrow G_\theta(q_0, [s]_k \cup M_i)$
      \IF{status == \texttt{False}}
        \STATE $[s]_N \leftarrow$ Retrieve top-$N$ Level-1 memories using $f_\theta(e_{q_i})$
        \STATE $[s]_m \leftarrow f_\gamma(q_i,[s]_N)$ \hfill // LLM retrieval
        \STATE $[p]_n \leftarrow$ Map $[s]_m$ to Level-2 memories
        \STATE $answer_i \leftarrow G_\gamma(q_0, [p]_n \cup M_i)$
      \ENDIF
      \STATE Update memory pool: $M_{i+1} \leftarrow answer_i$
      \STATE $q_{i+1}, \texttt{done} \leftarrow R(answer_i, q_0)$
      \IF{\texttt{done}}
        \STATE \textbf{Output:} $answer_i$
      \ENDIF
      \STATE $i \leftarrow i+1$
      \STATE $q_i \leftarrow q_{i+1}$
    \ENDWHILE
  \end{algorithmic}
\end{algorithm}

\onecolumn
\section{Detailed System Prompts}
\label{sec:appendix_prompt}
In this section, we provide a detailed overview of the prompts employed throughout the HyMem memory system. This includes the summary module in the storage phase, the generator $G_\theta$ in the Light Memory Module, the retriever $f_\gamma$ and generator $G_\gamma$ in the Deep Memory Module, the reflection module R , as well as the final model used for answer evaluation.

\subsection{Summary Module}
\begin{tcolorbox}[colback=gray!10, colframe=gray!80, boxrule=0.5pt, arc=3pt, left=2pt, right=2pt, top=2pt, bottom=2pt]
Your task is to extract all key information from the conversation and summarize each piece into a concise sentence. These sentences will serve as the memory content for subsequent agent responses.

The final output format should be:

\{ "keywords": ["Key information 1", "Key information 2", "Key information 3", ...] \}

Important instructions: 

All key information in the original conversation must be retained, including details such as time, location, persons involved, and significant events, to prevent errors in future responses due to missing details.

Key information should appropriately summarize the valuable content present in the original dialogue. To improve efficiency, all unnecessary dialogue elements (such as greetings, pleasantries, small talk, or casual remarks) must be excluded from the key information. 

Summaries should be concise and minimize character usage wherever possible. Whenever possible, integrate multiple related pieces of information into a single key sentence to increase information density. Avoid fragmenting key information excessively, as this could lead to unnecessary character usage.

\end{tcolorbox}
\subsection{Reflection $R$}

\begin{tcolorbox}[colback=gray!10, colframe=gray!80, boxrule=0.5pt, arc=3pt, left=2pt, right=2pt, top=2pt, bottom=2pt]
Your primary responsibility is to evaluate whether the current answer meets the standard based on the given question and the model's response.

If the answer is irrelevant to the question or contradicts the intent of the question, it should be judged as not meeting the standard. In such cases, set the finished field to 0, and rewrite the question by strengthening it based on what is missing in the answer, so that it can be used for further retrieval.

The newly generated question should be output in the new\_question field. If the answer is generally complete and well-reasoned, judge it as meeting the requirements and set the finished field to 1.
Here is an example output: \{ "finished": 0, "new\_question": "..." \}
\end{tcolorbox}
\subsection{Generator $G_\theta$}
\begin{tcolorbox}[colback=gray!10, colframe=gray!80, boxrule=0.5pt, arc=3pt, left=2pt, right=2pt, top=2pt, bottom=2pt]
You are a memory-based question answering assistant. You will receive a question along with a summary of related memory.

Your main responsibilities are as follows:

Based on the provided memory summary, determine whether the current question can be answered.

If the memory summary does not match the question, is incomplete, vague or ambiguous, is irrelevant, or if you are unsure whether you can answer, you must treat it as unanswerable. Note: Use strict criteria to avoid hallucinations and incorrect responses. In these cases, set the "finished" field to 2. More precise retrieval methods will be used later to provide more complete memory.

If a standard or golden answer to the question is clearly present in the memory summary, generate the answer in the "answer" field and set "finished" to 0. The answer must be complete and specific. For example, for time-related questions, to avoid ambiguity, clearly specify the reference point of any relative time expressions. If the answer is "last year", the correct format should be: "The current year is 2022, so the answer is last year."

The required output format is: \{"finished":0, "answer": "..."\}

Note: Return output only in JSON format. Do not provide any other form of output.
\end{tcolorbox}

\subsection{Retriever $f_\gamma$}

\begin{tcolorbox}[colback=gray!10, colframe=gray!80, boxrule=0.5pt, arc=3pt, left=2pt, right=2pt, top=2pt, bottom=2pt]
You are a text retriever.

You are given a question and a set of memory content indices. Each index is a brief summary of the key information in the corresponding memory content.

Your task is to, based on the given question, identify the ids of the memory indices that are most likely to provide context for answering the question.

Example:

Question: Where is Alice's home?

Indices:

id:0, dialogue time:13 October, 2022, Alice's two children

id:1, dialogue time:13 October, 2023, Alice's husband

id:2, dialogue time:23 October, 2022, Jack's job

id:3, dialogue time:13 October, 2022, Charity organization

id:4, dialogue time:31 October, 2022, Alice moved from her hometown

id:5, dialogue time:31 October, 2022, Alice's life in her hometown

Result:

\{ "keywords\_list": [4,5] \}
\end{tcolorbox}

\subsection{Generator $G_\gamma$}

\begin{tcolorbox}[colback=gray!10, colframe=gray!80, boxrule=0.5pt, arc=3pt, left=2pt, right=2pt, top=2pt, bottom=2pt]
You are a memory-based question-answering assistant. You will receive a question along with memories for answering it.

Your main responsibilities are as follows:

Please note that answers to questions are not always fixed. You should list all possible answers and strive to ensure the completeness of the information. For example, regarding time-related questions, to avoid ambiguity, explicitly specify the reference point for any relative time expressions. For instance, if the answer is "last year," the correct format should be: "The current year is 2022, and the answer is last year."
At the same time, not all questions have relevant memories, such as open-ended questions. Even if the memory information is incomplete, you can boldly infer the most likely answer based on the existing memories, even if the answer may be incomplete or not entirely rigorous. Avoid refusing to answer.

The required output format is: \{ "answer": "..." \}

Note: Only return your output in JSON format, and do not provide any other form of output. 
\end{tcolorbox}

\subsection{LLM as Judge}

\begin{tcolorbox}[colback=gray!10, colframe=gray!80, boxrule=0.5pt, arc=3pt, left=2pt, right=2pt, top=2pt, bottom=2pt]
You are an answer scoring expert.
You will receive the following information: (1) a question, (2) a standard (reference answer), (3) an answer generated by a memory-based LLM. Your task is to label the generated answer as CORRECT or WRONG.
Important notes:
Be as lenient as possible when scoring: the standard answer is typically a concise short sentence, while the generated answer may be longer and more detailed. As long as the generated answer aligns with the standard answer, it should be marked as CORRECT. Do not be overly nitpicky.
For time-related questions, the standard answer is a specific date, month, or year. The generated answer may be longer or use relative time expressions (e.g., "last Tuesday" or "next month"). Please be lenient in scoring—if the generated answer refers to the same time period as the standard answer, mark it as CORRECT.
Finally, provide CORRECT or WRONG.
Do not include both CORRECT and WRONG in your response, as this will cause errors in the evaluation script.
Simply return the label in JSON format with the key "label".
Now it's time for the formal question:
Question: \{question\}
Gold answer: \{gold\_answer\}
Generated answer: \{generated\_answer\}
\end{tcolorbox}

\end{document}